\documentclass[10pt,twocolumn,letterpaper]{article}

\usepackage[pagenumbers]{wacv} 

\usepackage[accsupp]{axessibility}  
\usepackage{graphicx}
\usepackage{amsmath}
\usepackage{amssymb}
\usepackage{booktabs}

\usepackage[utf8]{inputenc}
\usepackage{colortbl}
\usepackage{xcolor}
\usepackage{tcolorbox}
\usepackage[export]{adjustbox}
\usepackage{enumitem}
\usepackage{mathtools}
\usepackage{amssymb}
\usepackage{arydshln}

\usepackage{comment}
\usepackage{amsmath,amssymb} 
\usepackage{multirow}
\usepackage{adjustbox}
\usepackage[algo2e]{algorithm2e} 
\usepackage{algorithm}
\usepackage{algorithmic}
\usepackage{placeins}
\usepackage{bbm}
\SetKwRepeat{Do}{do}{while}

\usepackage[pagebackref,breaklinks,colorlinks]{hyperref}

\usepackage[capitalize]{cleveref}
\crefname{section}{Sec.}{Secs.}
\Crefname{section}{Section}{Sections}
\Crefname{table}{Table}{Tables}
\crefname{table}{Tab.}{Tabs.}

\def\wacvPaperID{1209} 
\def\confName{WACV}
\def\confYear{2024}

\begin{document}

\title{Training-Based Model Refinement and Representation Disagreement for Semi-Supervised Object Detection}

\author{Seyed Mojtaba Marvasti-Zadeh\\
University of Alberta\\
{\tt\small seyedmoj@ualberta.ca}
\and
Nilanjan Ray\\
University of Alberta\\
{\tt\small nray1@ualberta.ca}
\and
Nadir Erbilgin\\
University of Alberta\\
{\tt\small erbilgin@ualberta.ca}
}
\maketitle
\begin{abstract}
    Semi-supervised object detection (SSOD) aims to improve the performance and generalization of existing object detectors by utilizing limited labeled data and extensive unlabeled data. Despite many advances, recent SSOD methods are still challenged by inadequate model refinement using the classical exponential moving average (EMA) strategy, the consensus of Teacher-Student models in the latter stages of training (i.e., losing their distinctiveness), and noisy/misleading pseudo-labels. 
    This paper proposes a novel training-based model refinement (TMR) stage and a simple yet effective representation disagreement (RD) strategy to address the limitations of classical EMA and the consensus problem. The TMR stage of Teacher-Student models optimizes the lightweight scaling operation to refine the model's weights and prevent overfitting or forgetting learned patterns from unlabeled data. Meanwhile, the RD strategy helps keep these models diverged to encourage the student model to explore additional patterns in unlabeled data. 
    Our approach can be integrated into established SSOD methods and is empirically validated using two baseline methods, with and without cascade regression, to generate more reliable pseudo-labels.    
    Extensive experiments demonstrate the superior performance of our approach over state-of-the-art SSOD methods. Specifically, the proposed approach outperforms the baseline Unbiased-Teacher-v2 (\& Unbiased-Teacher-v1) method by an average mAP margin of $2.23$, $2.1$, and $3.36$ (\& $2.07$, $1.9$ and $3.27$) on \texttt{COCO-standard}, \texttt{COCO-additional}, and \texttt{Pascal VOC} datasets, respectively.
\end{abstract}
\begin{figure}[t!]
\centering
\includegraphics[width=0.8\linewidth, height=5.25cm, valign=c]{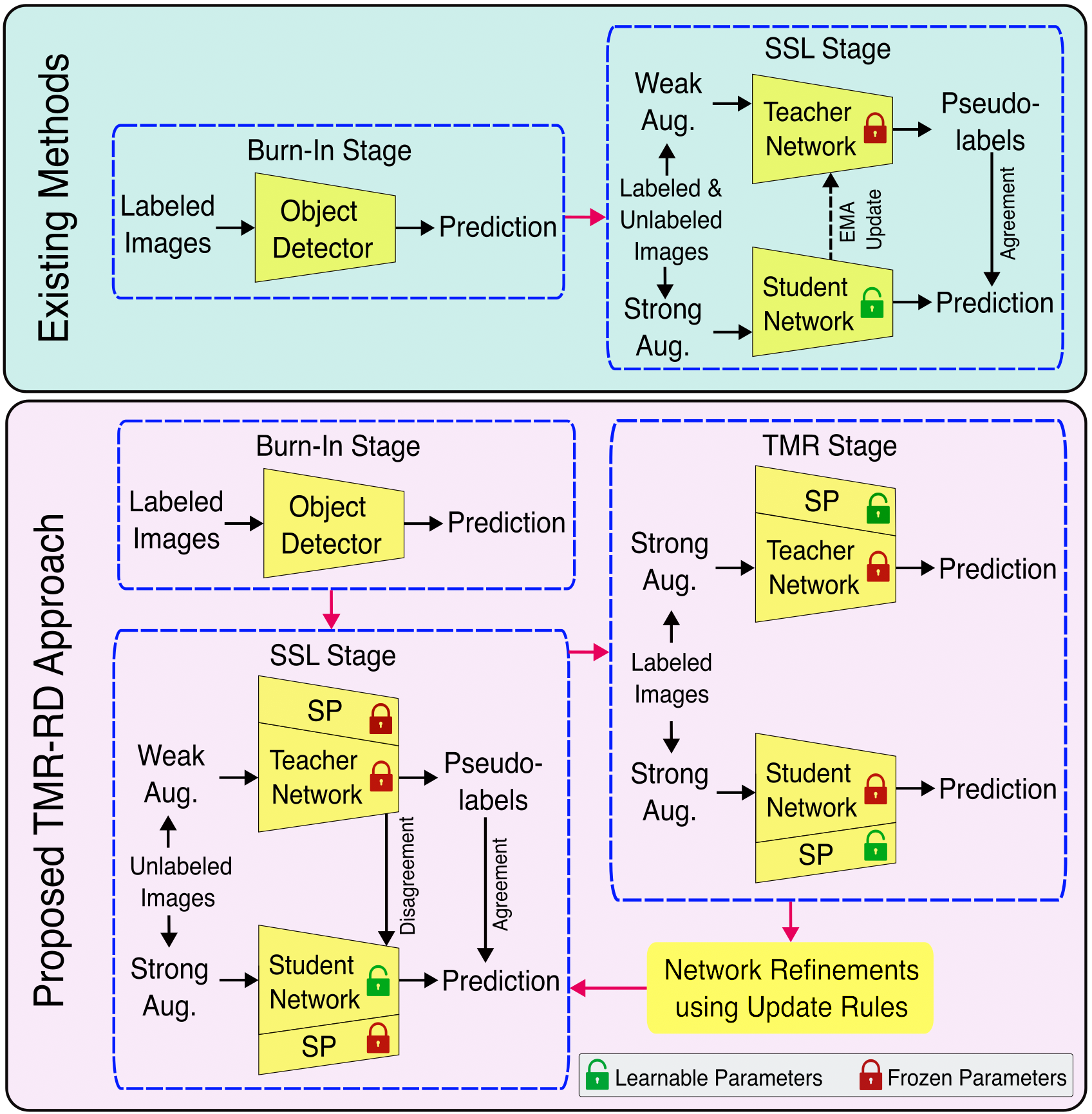}
\vskip -.2cm
\caption{\small An illustration of the difference between existing SSOD methods and the proposed approach. Existing methods use Teacher-Student models (initialized in the Burn-In stage) where the student model is trained with pseudo-labels generated by the teacher model, and teacher model weights are updated using the classical EMA strategy. In the proposed approach, during the SSL stage, the teacher model weights and scaling operation parameters (denoted as SPs) remain frozen during the training of the student model. After completing $N$ iterations, the weights of the models are frozen, and the TMR stage optimizes lightweight SPs for $N'$ iterations, followed by adaptive refinement of model weights using updating rules. The SSL and TMR stages are iteratively applied until the completion of the training process. Best viewed in color.
} \label{fig:overview}
\vskip -0.5cm
\end{figure}
\section{Introduction}
\label{sec:intro}
Object detection has experienced significant progress owing to the availability of large-scale benchmark datasets containing pairs of class labels and bounding boxes for various objects within images. However, collecting and accurately annotating object detection datasets is extremely expensive, time-consuming, and labor-intensive due to the lack of domain experts, limited resources, and the complicated nature of the problem. Meanwhile, acquiring a large amount of unlabeled data is relatively easy and provides valuable insights into the data distribution from which robust representations (under various transformations) can be learned \cite{SmallDataEra_Survey}. 
\begin{table*}[t]
\captionsetup{font=small}
\caption{
Comparative review of SSOD methods. Best viewed with zoom-in.
} 
\vskip -.3cm
\centering 
\resizebox{\textwidth}{!}{
\begin{tabular}{c c c c c c c}  
\hline \hline
Method & Motivation(s) & Backbone(s) & Detector(s) & Teacher Update & Weak Augmentation(s) & Strong Augmentation(s) \\ \hline 

 \cellcolor{teal!10}CSD \cite{CSD} & \cellcolor{teal!10}Efficient training process (than self-training) & \cellcolor{teal!10}ResNet-101 & \cellcolor{teal!10}SSD, RFCN & \cellcolor{teal!10}$\times$ & \cellcolor{teal!10}Horizontal flip & \cellcolor{teal!10}$\times$         \\

 STAC \cite{STAC} & Combining self-training \& consistency regularization & ResNet-50-FPN  & Faster-RCNN & $\times$ & $\times$ & Color/geometric transformations, cutout  \\

 \cellcolor{teal!10}ISMT \cite{ISMT} & \cellcolor{teal!10} Detecting training iteration discrepancies & \cellcolor{teal!10}ResNet-50-FPN & \cellcolor{teal!10}Faster-RCNN & \cellcolor{teal!10}Classical EMA & \cellcolor{teal!10}$\times$ & \cellcolor{teal!10}Color jitter \\

 Soft-Teacher \cite{SoftTeacher} & More reliable pseudo-labels & ResNet-50-FPN & Faster-RCNN & Classical EMA & Horizontal flip  & Scale/solarize/brightness/contrast/sharpness jitters, translation, rotate, shift, cutout \\

 \cellcolor{teal!10}Humble-Teacher \cite{HumbleTeacher} & \cellcolor{teal!10}More reliable pseudo-labels, Teacher ensemble & \cellcolor{teal!10}ResNet-50-FPN, ResNet-152-FPN  & \cellcolor{teal!10}Faster-RCNN, Cascade-RCNN & \cellcolor{teal!10}Classical EMA & \cellcolor{teal!10}Resize, flip & \cellcolor{teal!10}color/sharpness/contrast jitters, Gaussian noise, cutout  \\

 Instant-Teaching \cite{InstantTeaching} & More reliable pseudo-labels, Confirmation bias & ResNet-50-FPN & Faster-RCNN & Co-rectify scheme & $\times$ &  Color/geometric transformations, cutout, mixup, mosaic \\

 \cellcolor{teal!10}Unbiased-Teacher-v1 \cite{UnbiasedTeacher} & \cellcolor{teal!10}Class imbalance, Pseudo-label bias & \cellcolor{teal!10}ResNet-50-FPN  & \cellcolor{teal!10}Faster-RCNN & \cellcolor{teal!10}Classical EMA & \cellcolor{teal!10}Horizontal flip & \cellcolor{teal!10}Color jitter, grayscale, Gaussian blur, cutout \\

 Active-Teacher \cite{ActiveTeacher} & More reliable pseudo-labels, Data initialization & ResNet-50-FPN & Faster-RCNN & Classical EMA & Horizontal flip & Horizontal flip, color jitter, grayscale, Gaussian blur, cutout  \\

 \cellcolor{teal!10}ACRST \cite{ACRST} & \cellcolor{teal!10}Class imbalance, Biased/noisy pseudo-labels & \cellcolor{teal!10}ResNet-50-FPN & \cellcolor{teal!10}Faster-RCNN & \cellcolor{teal!10}Classical EMA & \cellcolor{teal!10}Resize, flip, crop & \cellcolor{teal!10}Color jitter, Gaussian blur, cutout \\

 CAPL \cite{CertaintyPseudo} & Class imbalance, Localization precision & ResNet-50-FPN & Faster-RCNN & $\times$ & Horizontal flipping  & Color jitter, Gaussian blur, cutout \\

\cellcolor{teal!10}SED \cite{SED} & \cellcolor{teal!10}Class imbalance, Large object size variance & \cellcolor{teal!10}ResNet-50-FPN & \cellcolor{teal!10}Faster-RCNN & \cellcolor{teal!10}Classical EMA & \cellcolor{teal!10}Resize, horizontal flip & \cellcolor{teal!10}Color jitter, grayscale, Gaussian blur, cutout \\

Label-Match \cite{LabelMatch} & Label mismatch, Confirmation bias & ResNet-50-FPN & Faster-RCNN & Classical EMA & Horizontal flip, multi-scale & Color jitter, grayscale, Gaussian blur, cutout \\

\cellcolor{teal!10}MA-GCP \cite{MA-GCP} & \cellcolor{teal!10}Relation between labeled \& unlabeled data & \cellcolor{teal!10}ResNet-50-FPN & \cellcolor{teal!10}Faster-RCNN & \cellcolor{teal!10}Classical EMA & \cellcolor{teal!10}Horizontal flip & \cellcolor{teal!10}Scale/solarize/brightness/contrast/sharpness jitters, translation, rotate, shift, cutout \\

MUM \cite{MUM} & More complex data augmentation & ResNet-50-FPN, SwinTransformer & Faster-RCNN & Classical EMA & Horizontal flip  & Mixing image tiles, color jitter, grayscale, Gaussian blur, cutout \\

 \cellcolor{teal!10}Unbiased-Teacher-v2 \cite{UnbiasedTeacher2} & \cellcolor{teal!10}Anchor-free detectors, \cellcolor{teal!10}Misleading pseudo-labels & \cellcolor{teal!10}ResNet-50-FPN & \cellcolor{teal!10}Faster-RCNN, FCOS & \cellcolor{teal!10}Classical EMA & \cellcolor{teal!10}Horizontal flip & \cellcolor{teal!10}Color/scale jitters, grayscale, Gaussian blur, cutout \\

Diverse-Learner \cite{DiverseLearner} & Maintain networks distinctiveness & ResNet-50-FPN & Faster-RCNN & Classical EMA & Horizontal flip, random size & Random erasing, rotation, color jitters, etc. \\

\cellcolor{teal!10}PseCo \cite{PseCo} & \cellcolor{teal!10}Noisy pseudo-labels, Scale-invariant learning & \cellcolor{teal!10}ResNet-50-FPN & \cellcolor{teal!10}Faster-RCNN & \cellcolor{teal!10}Classical EMA & \cellcolor{teal!10}Horizontal flip & \cellcolor{teal!10}Scale/solarize/brightness/contrast/sharpness jitters, translation, rotate, shift, cutout \\

VC-Learner \cite{VCLearning} & Confirmation bias, Confusing samples & ResNet-50-FPN & Faster-RCNN & $\times$ & Horizontal flip & Color/scale jitters, grayscale, Gaussian blur, cutout \\

\cellcolor{teal!10}De-biased Teacher \cite{De-biased-Teacher} & \cellcolor{teal!10}IoU matching bias & \cellcolor{teal!10}ResNet-50-FPN & \cellcolor{teal!10}Faster-RCNN & \cellcolor{teal!10}Classical EMA & \cellcolor{teal!10}Horizontal flip & \cellcolor{teal!10}Color jitter, grayscale, Gaussian blur, cutout \\

Pseudo-Polish \cite{Pseudo-Polish} & Limited teacher model generalization & ResNet-50-FPN & Faster-RCNN & Classical EMA & Horizontal flip  & Scale/solarize/brightness/contrast/sharpness jitters, translation, rotate, shift, cutout \\ 
\hline \hline
\end{tabular}
\label{table:comp_ssod}
}
\vskip -.4cm
\end{table*}
Hence, leveraging unlabeled data for training object detectors has become increasingly popular due to its potential to significantly reduce annotation costs \& efforts while improving model performance and generalization. \textit{Semi-supervised object detection} (SSOD) aims to harness unsupervised information in scenarios where labeled data is limited. Most recent SSOD methods rely on self-training techniques, in which the pseudo-labels generated from the teacher model(s) are utilized to train the student model(s) when weak \& strong augmentations of images are used as inputs (see Fig.~\ref{fig:overview} \& Table~\ref{table:comp_ssod}). While these methods enforce the consistency (or agreement) of Teacher-Student predictions to train the student model, the classical \textit{exponential moving average} (EMA) strategy is adopted to evolve the teacher model progressively. 
This strategy gradually refines the weights of the teacher model to improve the accuracy of pseudo-labels \& resiliency to noisy weights of the student model \cite{HumbleTeacher} as well as alleviate the adverse effects of pseudo-labeling bias \cite{UnbiasedTeacher}. \\
\indent Despite the progress made in the SSOD, training Teacher-Student models still faces three major challenges. 
The first challenge pertains to applying the classical EMA strategy using manually defined smoothing coefficients. It can lead to two potential issues when refining the weights of the teacher model: 
i) insensitivity to important changes in weights of the student model due to excessive reliance on the initialized detector, and 
ii) sub-optimal performance because constant coefficients may not be effective for all refinement steps, resulting in weaker pseudo-labels. 
To address these limitations, we propose a novel \textit{training-based model refinement} (TMR) stage that adaptively refines the weights of Teacher-Student models (see Fig.~\ref{fig:overview}). This stage is added to the commonly used training stages of i) pre-training on limited labeled data (or Burn-In stage), and ii) \textit{semi-supervised learning} (SSL) stage in SSOD methods. Inspired by \textit{meta-transfer learning} (MTL) \cite{MTL}, we optimize the lightweight scaling operation corresponding to learnable parameters of Teacher-Student models to effectively aggregate information from labeled and unlabeled data.
The models' weights can then be dynamically refined using the introduced update rules to ensure the teacher model is up-to-date and reduce the effect of noisy pseudo-labels on the student model without the risk of overfitting or forgetting the patterns learned from unlabeled data. \\
\indent The second challenge is the consensus problem (i.e., losing the distinctiveness of Teacher-Student models) at the latter stages of the training procedure when two models become almost identical. This is also derived from the classical EMA strategy, leading to teacher weights being close to student ones as training progresses. Accordingly, both models generate similar predictions and make it difficult for the teacher model to extract helpful information from unlabeled data for supervising the student model. 
To alleviate this issue, we propose a simple yet effective \textit{representation disagreement} (RD) strategy that incorporates the asymmetric \textit{Kullback-Leiber} (KL) divergence between the semantic representation of Teacher-Student models to prevent early convergence. 
This strategy aims to increase model divergence by encouraging the student model to explore more robust representations, learn complementary information, and reduce the memorization effect of easy samples. \\
\indent The last challenge is to address noisy/misleading pseudo-labels, which can impede accurate model optimization leading to ineffective learning from unlabeled data and slow convergence. As shown in Table~\ref{table:comp_ssod}, extensive efforts have been made to provide more reliable pseudo-labels. Although most of these methods rely on the well-established Faster-RCNN detector \cite{FasterRCNN}, it can be optimal solely for detecting objects at a single-quality level due to the adversarial nature of producing noisy boxes or assembling inadequate positive proposals. 
Following the Humble-Teacher \cite{HumbleTeacher}, we integrate cascade regression into our baselines (i.e., \cite{UnbiasedTeacher, UnbiasedTeacher2}) to generate more reliable pseudo-labels while reducing the overfitting problem. 
Note that we consider this integration \textit{{not as our contribution}} but rather as a means to highlight the versatility of our proposed TMR-RD approach with various base detectors.
To conduct a fair comparison, our empirical experiments (Sec.~\ref{sec4:exp}) involve using two SSOD baselines with or without cascade regression. \\
\indent The main contributions are summarized as follows: 
(1) A novel TMR stage is proposed to dynamically refine the weights of Teacher/Student models in SSOD frameworks. It can learn the lightweight scaling operation corresponding to the learnable parameters of the Teacher-Student models, enabling fast convergence without the risk of overfitting or forgetting patterns learned from unlabeled data. To the best of our knowledge, this is the first work on the dynamic refinement of Teacher-Student learning, with the potential to inspire further research in its applications (e.g., knowledge exchange, domain adaptation, etc.).
(2) A simple yet effective RD strategy is proposed to alleviate the consensus problem of Teacher-Student models with progress in training. This strategy prevents the models from converging too early, allowing better generalization through more exploring underlying patterns in unlabeled data.
(3) Extensive empirical evaluations and ablation analyses demonstrate superior performance and generalizability of the proposed approach, which can be incorporated into existing SSOD methods to boost their performance. 
\section{Related Work} \label{sec2:relatedwork}
\subsection{Classical Exponential Moving Average (EMA)} \label{sec:ema}
While existing SSOD methods tackle various challenges (e.g., efficient training \cite{CSD, STAC}, detection discrepancies \cite{ISMT}, localization certainty \cite{CertaintyPseudo}, prediction consistency \cite{MA-GCP}, object size \cite{SED}, data augmentation \cite{MUM}, confusing samples \cite{VCLearning}, and reliability of pseudo-labels \cite{SoftTeacher, InstantTeaching, HumbleTeacher, ActiveTeacher}), they mostly employ the classical EMA strategy \cite{EMA_MeanTeachers} to refine the teacher's model weights (see Table~\ref{table:comp_ssod}).
This strategy updates this model by averaging its weights to ensure more accurate predictions from the teacher model than the student one. It can minimize the adverse effects of imbalanced and noisy labels through mutually reinforcing pseudo-labeling and the detection training steps by
\vskip -.5cm
\begin{equation} \label{eq:ema}
\begin{aligned}
\theta_{t}^{n} &\leftarrow \alpha \theta_{t}^{n-1} + (1-\alpha)\theta_{s}^{n-1} \\
\Rightarrow \quad \theta_{t}^{n} &\leftarrow \alpha^{n} \theta_{t}^{0} + (1-\alpha)\sum_{k=0}^{n-1}\alpha^{n-1-k}\theta_{s}^{k}
\end{aligned}
\end{equation}
\vskip -.3cm
\noindent in which $\theta_{t}^{0}$, $\theta_{t}^{n}$, and $\theta_{s}^{n-1}$ represent the initialized detector (trained on limited labeled data) at time step ${n=0}$, the parameters of the teacher model at $n$-th time step, and the parameters of the student model at $(n-1)$-th time step, respectively. In addition, $\alpha$ is a manually-defined smoothing coefficient (i.e., EMA decay) often set to $0.999$ in SSOD methods so that the teacher model can benefit from a long memory while it assumes the student model improves slowly. 
The limitations of this strategy (also presented in Sec.~\ref{sec:intro}) have recently been discussed in the Diverse-Learner \cite{DiverseLearner}, which uses two Teacher-Student pairs and multi-threshold classification loss to alleviate the associated drawbacks, maintain distinctiveness between models, and improve pseudo-labels. However, it still uses the classical EMA strategy to refine its teacher models. 
In this paper, we propose a novel approach for refining Teacher-Student models by optimizing the lightweight scaling operation corresponding to the learnable model parameters, addressing the limitations of the classical EMA strategy.

\begin{figure*}[t!]
\centering
\includegraphics[width=0.88\linewidth,valign=c]{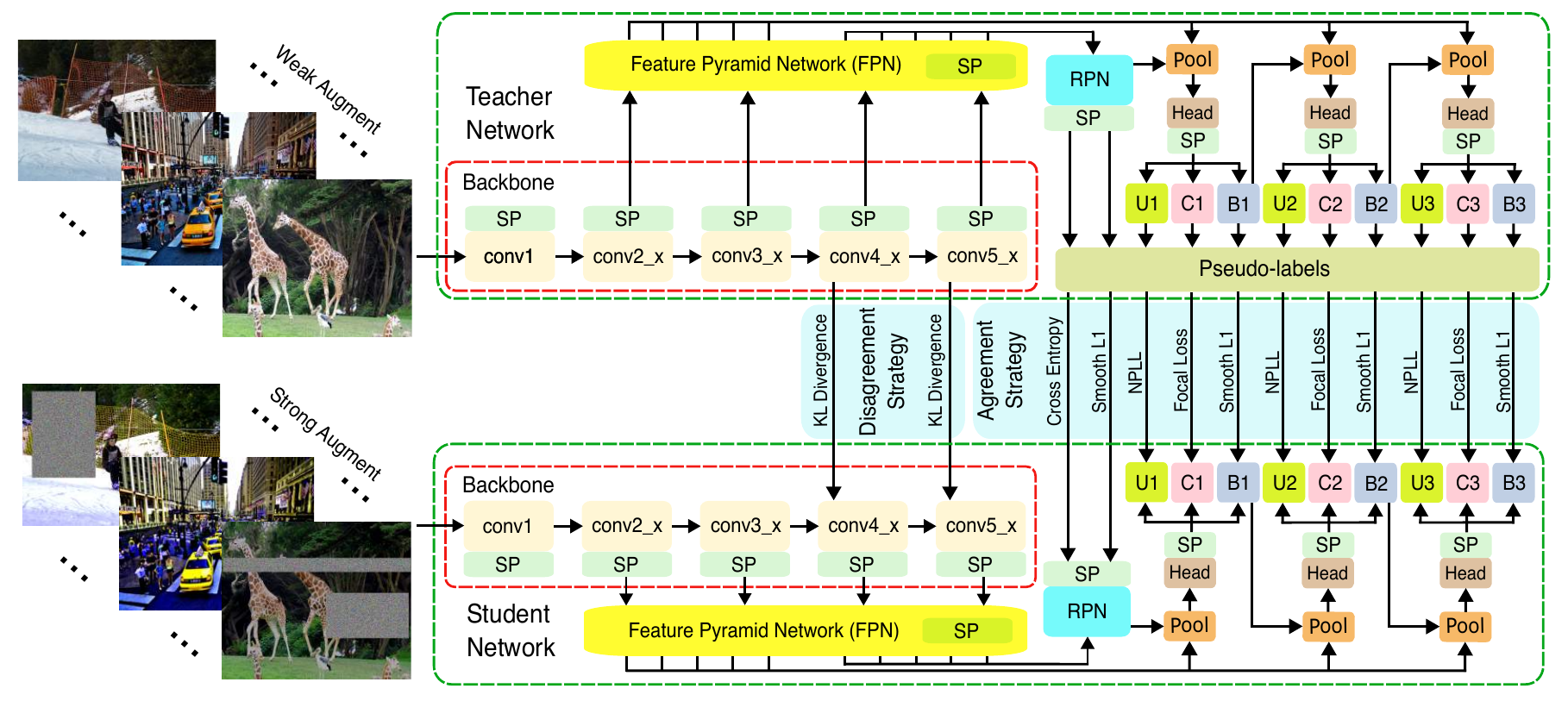}
\vskip -.2cm
\caption{\small Overview of integrating the TMR-RD approach and cascade regression into the baseline Unbiased-Teacher-v2 \cite{UnbiasedTeacher2}. SPs, U, C, B, and NPLL represent scaling operation parameters, localization uncertainty, ROI classification, boundary distance, and negative power log-likelihood loss, respectively. For comprehensive and fair comparisons, we also evaluate our models w/o the uncertainty prediction head (e.g., Unbiased-Teacher-v1 \cite{UnbiasedTeacher}) and w/o cascade regression (i.e., Faster-RCNN).
} \label{fig:TNR-FD}
\vskip -0.3cm
\end{figure*}
\subsection{Model Robustness with Reliable Pseudo-labels} \label{sec:ssod}
Recent advancements have been primarily directed at enhancing the reliability of pseudo-labels (see Table~\ref{table:comp_ssod}). These improvements encompass several approaches, such as:
i) selecting pseudo-boxes with higher scores \cite{SoftTeacher},
ii) employing instant pseudo-labeling and simultaneous training of two models \cite{InstantTeaching},
iii) utilizing soft pseudo-labels and teacher ensembles \cite{HumbleTeacher},
iv) selecting the most optimal labeled examples \cite{ActiveTeacher},
v) employing a memory module and a two-stage pseudo-label filtering \cite{ACRST},
vi) using a redistributed mean teacher and proposal self-assignment \cite{LabelMatch},
vii) applying prediction-guided label assignment and positive-proposal consistency voting \cite{PseCo},
viii) utilizing Teacher-Student mutual learning \cite{UnbiasedTeacher},
ix) assessing relative model uncertainties \cite{UnbiasedTeacher2},
x) developing two differently structured pseudo-label polishing networks \cite{Pseudo-Polish}, and
xi) directly generating proposals for consistency regularization between perturbed image pairs \cite{De-biased-Teacher}.
These methods aim to train robust models by imposing an agreement strategy (or consistency) between the predictions of Teacher-Student models using labeled and unlabeled data. However, relying on similar predictions and classical EMA can lead to losing model distinctiveness and early convergence of models. To address this, we introduce a simple yet effective RD strategy, which helps maintain divergence between the models, enabling the exploration of additional patterns in unlabeled data and promoting model generalization. 

Aside from that, existing methods typically use a confidence threshold to determine the number of bounding boxes. Decreasing this threshold increases the information mined but can introduce noisy pseudo-labels, reducing detection performance. Conversely, using a higher threshold produces a limited number of high-quality boxes, ignoring some objects during the SSL stage. 
In addition to the Faster-RCNN detector, we assess our TMR-RD approach with the Faster-RCNN equipped with cascade regression, which relies on a multi-stage architecture featuring specialized regressors for precise localization. This evaluation serves to highlight the versatility of our approach, indicating its potential to improve the overall performance and generalization of existing methods. 
\section{Proposed Approach: TMR-RD} \label{sec3:ours}
In this section, we present the TMR-RD approach, which can be integrated into most existing SSOD methods employing Teacher-Student models and the classical EMA strategy. 
We utilize the well-designed Unbiased-Teacher-v1 \cite{UnbiasedTeacher} and Unbiased-Teacher-v2 \cite{UnbiasedTeacher2}, both with and without cascade regression, as the baseline SSOD methods to demonstrate our approach's effectiveness.
For the sake of generality, we will describe the integration of our approach into the Unbiased-Teacher-v2 (named TMR-RD-v2) with the base detector using cascade regression, as shown in Fig.~\ref{fig:TNR-FD}. Still, our approach can be readily adapted to the Unbiased-Teacher-v1 without the additional branch of localization uncertainty (named TMR-RD-v1) and work with the Faster-RCNN detector (single-stage regression). 
An overview of our approach is shown in Algorithm~\ref{algorithm}.
The experimental results for various models are provided in Sec.~\ref{sec4:exp}.
\subsection{Training-based Model Refinement (TMR)} \label{subsec:update}
\vskip -.1cm
The proposed TMR stage is inspired by the MTL \cite{MTL} that adapts large-scale trained base classification models to new tasks with limited data using two lightweight neuron operations of scaling and shifting.
However, the proposed TMR is designed following the EMA equation (see Eq.~\ref{eq:ema}) that merely requires scaling coefficients to refine the model weights. 
The proposed approach comprises three stages of Burn-In, SSL, and TMR (see Fig.~\ref{fig:overview}), of which the first is the same as that in the baseline methods \cite{UnbiasedTeacher, UnbiasedTeacher2}. The Burn-In stage provides a good initialization by training the base object detector on the available labeled data.
Then, the initialized detector is duplicated into the Teacher-Student models so that pseudo-labels are generated by the Teacher and used to train the Student during the SSL stage. However, the weights of the Teacher are not updated using the classical EMA. Instead, we freeze neuron weights from the SSL stage after completing $N$ iterations and initiate the TMR stage to learn a set of lightweight scaling operations associated with the trained parameters of the Teacher-Student models. 
By completing $N'$ iterations, the proposed TMR can promote the progressive transfer of knowledge from the continually learning student model to the teacher one, thereby aggregating information more effectively and improving the generation of pseudo-labels.
\\
\indent In the TMR stage, the scaling operation is denoted as $\Omega_{i}$, where $i\in\left(t, s\right)$ refers to the teacher or student model. The MTL \cite{MTL} defines the scaling and shifting operations exclusively for the feature extractor (i.e., backbone layers) and updates them after optimizing a temporal classifier (i.e., as per a new task). However, we apply scaling operations to all frozen neuron weights (see Fig.~\ref{fig:TNR-FD}) of Teacher/Student by $\mathcal{S}(\hat{x};\theta_{i};\Omega_{i})=(\theta_{i}\odot \Omega_{i})\hat{x}_{k}$, where $\mathcal{\hat{D}}=\{\hat{x}_k, \hat{y}_k\}_{k=1}^{N_{sup}}$ and $\odot$ are the strongly-augmented labeled data and the element-wise multiplication, respectively.
Then, the TMR loss is defined as
\vskip -0.6cm
\begin{gather} \label{eq:L_scale}
\mathcal{L}_{\rm{TMR}}= \lambda_t \mathcal{L}_{sc}\left ([\theta_{t}; \Omega_{t}] \right) + 
\lambda_s \mathcal{L}_{sc}\left ([\theta_{s}; \Omega_{s}] \right),
\end{gather}
\vskip -0.7cm
\begin{equation} \label{eq:L_cascade}
\begin{aligned}
&\mathcal{L}_{sc}\left ([\theta_{i}; \Omega_{i}] \right) = \sum\nolimits_{j\in\mathcal{\hat{D}}}
\mathcal{L}_{cls}^{rpn}\left(\mathcal{C}_{[\theta_{i}; \Omega_{i}]}^{0}({\hat{x}}_{j}^{0}), {\hat{y}}_{j}^{0}\right) \\ 
&+ \mathcal{L}_{reg}^{rpn}\left(\mathcal{B}_{[\theta_{i}; \Omega_{i}]}^{0}({\hat{x}}_{j}^{0}), {\hat{y}}_{j}^{0}\right) + 
\mathcal{L}_{cls}^{roi}\left(\mathcal{C}_{[\theta_{i}; \Omega_{i}]}^{1}({\hat{x}}_{j}^{1}), {\hat{y}}_{j}^{1}\right)  \\ 
&+ \mathcal{L}_{reg}^{roi}\left(\mathcal{B}_{[\theta_{i}; \Omega_{i}]}^{1}({\hat{x}}_{j}^{1}), {\hat{y}}_{j}^{1}\right) + 
\mathcal{L}_{reg}^{roi}\left(\mathcal{U}_{[\theta_{i}; \Omega_{i}]}^{1}({\hat{x}}_{j}^{1}), {\hat{y}}_{j}^{1}\right)  \\ 
&+ \mathcal{L}_{cls}^{roi}\left(\mathcal{C}_{[\theta_{i}; \Omega_{i}]}^{2}({\hat{x}}_{j}^{2}), {\hat{y}}_{j}^{2}\right) + 
\mathcal{L}_{reg}^{roi}\left(\mathcal{B}_{[\theta_{i}; \Omega_{i}]}^{2}({\hat{x}}_{j}^{2}), {\hat{y}}_{j}^{2}\right)  \\ 
&+ \mathcal{L}_{reg}^{roi}\left(\mathcal{U}_{[\theta_{i}; \Omega_{i}]}^{2}({\hat{x}}_{j}^{2}), {\hat{y}}_{j}^{2}\right) + 
\mathcal{L}_{cls}^{roi}\left(\mathcal{C}_{[\theta_{i}; \Omega_{i}]}^{3}({\hat{x}}_{j}^{3}), {\hat{y}}_{j}^{3}\right)  \\ 
&+ \mathcal{L}_{reg}^{roi}\left(\mathcal{B}_{[\theta_{i}; \Omega_{i}]}^{3}({\hat{x}}_{j}^{3}), {\hat{y}}_{j}^{3}\right) +
\mathcal{L}_{reg}^{roi}\left(\mathcal{U}_{[\theta_{i}; \Omega_{i}]}^{3}({\hat{x}}_{j}^{3}), {\hat{y}}_{j}^{3}\right),
\end{aligned}
\end{equation}
\noindent where $\mathcal{L}_{sc}$, $\mathcal{C}^{k}$, $\mathcal{B}^{k}$, and $\mathcal{U}^{k}$ represent the scaling operation loss, a classifier at stage $k$, a boundary regressor at stage $k$, and localization uncertainty at stage $k$, respectively. Each regression stage is optimized for an IoU threshold $\tau^{k}$, respectively. In addition, $rpn$ and $roi$ refer to the RPN and RoI-Head branches, respectively.
As shown in Fig.~\ref{fig:TNR-FD}, we utilize the Cross-Entropy, Smooth-L1, Focal loss, and \textit{negative power log-likelihood loss} (NPLL) as in the baseline methods \cite{UnbiasedTeacher, UnbiasedTeacher2} for our models to provide fair comparisons.
Next, the scaling operation weights associated with learnable parameters of the Teacher/Student are updated by 
\vskip -0.6cm
\begin{gather} \label{eq:scale_upd}
\Omega_{i} =: \Omega_{i} - \gamma \nabla_{\Omega_{i}}\mathcal{L}_{\rm{TMR}}, 
\end{gather}
\vskip -0.2cm
\noindent in which $\gamma$ denotes the learning rate. 
Optimizing the lightweight scaling operation while keeping the large-scale trained weights of Teacher-Student models unchanged allows for fast convergence while reducing the overfitting risk. 
After $N'$ iterations, the models are refined using our updating rules as
\vskip -.6cm
\begin{gather} \label{eq:rules}
\theta_{t}^{n}=\frac{\Omega_{t}}{\Omega_{t}+\Omega_{s}}\theta_{t}^{n-1} + \frac{\Omega_{s}}{\Omega_{t}+\Omega_{s}}\theta_{s}^{n-1}, \\
\label{eq:rules_stu}
\theta_{s}^{n}=(1-\frac{\Omega_{t}}{\Omega_{t}+\Omega_{s}})\theta_{t}^{n-1} + (1-\frac{\Omega_{s}}{\Omega_{t}+\Omega_{s}})\theta_{s}^{n-1}.        
\end{gather}
\vskip -0.1cm
\noindent These rules imply that the scaling operation weights stay within the permissible range of zero to one and selectively refine models by the ability to adjust or forget inaccurate model weights more effectively.
Besides updating the teacher weights, the student weights are also slightly refined to reduce the impact of noisy pseudo-labels from training this model with potentially misleading pseudo-labels in the SSL stage. The SSL and TMR stages are alternatively continued until the end of the training procedure.
This approach makes it possible to efficiently learn EMA coefficients on-the-fly, improving the performance of SSOD methods.
\begin{figure}[!b]
\vspace{-.7cm} 
\begin{algorithm}[H]
\algsetup{linenosize=\tiny}
\caption{\textbf{:} Proposed TMR-RD Approach}
\label{algorithm}
{\textsc{\textbf{\scriptsize Input:}}} \scriptsize Labeled images $\mathcal{D}_{\rm{sup}}$, Unlabeled images $\mathcal{D}_{\rm{unsup}}$, Detection networks (w/o cascade reg.): Teacher $\mathbbm{Net}(\cdot;{\theta_t})$ \& Student $\mathbbm{Net}(\cdot;{\theta_s})$ \\
{\textsc{\textbf{Output:}}} Trained $\mathbbm{Net}(\cdot;{\theta_t})$ \& $\mathbbm{Net}(\cdot;{\theta_s})$
\\
\vskip -.4cm
\centerline{\rule{7cm}{0.1pt}} 
\scriptsize
Freeze $\Omega_{t}$ \hspace{5.35 cm} \textcolor{teal}{$\triangleright$ {\fontsize{6}{8}\selectfont \texttt{Burn-In Stage}}} \\
\For{samples in $\mathcal{D}_{\rm{sup}}$}{ 
    Evaluate $\mathcal{L}_{\rm{sup}}$ following \cite{UnbiasedTeacher,UnbiasedTeacher2} \\
    Optimize $\theta_t$
}
Initialize $\mathbbm{Net}(\cdot;{\theta_s})$ by setting $\theta_s$ $\leftarrow$ $\theta_t$ \\
\While{not done}{
    Freeze $\Omega_{t}$, $\Omega_{s}$, and $\theta_t$ \hspace{4.1 cm} \textcolor{teal}{$\triangleright$ {\fontsize{6}{8}\selectfont \texttt{SSL Stage}}} \\
    \For{$N$ iterations}{
    Samples from $\bar{\mathcal{D}}_{\rm{unsup}}$ and $\hat{\mathcal{D}}_{\rm{unsup}}$ \hspace{1.1159 cm} \textcolor{teal}{$\triangleright$ {\fontsize{6}{8}\selectfont \texttt{Weak \& Strong Aug.}}} \\
    Evaluate $\mathcal{L}_{\rm{unsup}}$ as in \cite{UnbiasedTeacher2} for TMR-RD-v2 (or \cite{UnbiasedTeacher} for TMR-RD-v1) \\
    Evaluate $\mathcal{L}_{\rm{RD}}$ using Eq.~\ref{eq:kl_loss} \\
    Optimize $\theta_s$ using Eq.~\ref{eq:stu_upd_kl}
    }
    Freeze $\theta_t$ and $\theta_s$ \hspace{4.6 cm} \textcolor{teal}{$\triangleright$ {\fontsize{6}{8}\selectfont \texttt{TMR Stage}}} \\
    \For{$N'$ iterations}{
    Samples from $\hat{\mathcal{D}}_{\rm{sup}}$ \hspace{3.57 cm} \textcolor{teal}{$\triangleright$ {\fontsize{6}{8}\selectfont \texttt{Strong Aug.}}} \\
    Evaluate $\mathcal{L}_{\rm{TMR}}$ using Eq.~\ref{eq:L_scale} (single-stage regression of Eq.~\ref{eq:L_cascade}) \\
    Optimize $\Omega_{t}$ and $\Omega_{s}$ using Eq.~\ref{eq:scale_upd}
    }
    Refine $\theta_t$ and $\theta_s$ using Eq.~\ref{eq:rules} and Eq.~\ref{eq:rules_stu}
}
\end{algorithm}
\vspace{-.5cm} 
\end{figure}
\subsection{Representation Disagreement (RD)} \label{subsec:disagree}
The disagreement strategy idea lies in the principles of Co-training \cite{Disagree_idea}, where the effectiveness of an ensemble can be improved by keeping divergent classifiers. It has then been extended to train deep networks in the presence of noisy labels (e.g., \cite{Decoupling, CoTeachingPlus}). These scenarios involve simultaneous training of two deep networks based on the cross-update principle implied by the culture-evolving hypothesis \cite{EvolveBengio}, where a network can improve its learning capability when it is aided by signals generated by another network. Although two networks with distinct learning capabilities can distinguish different error categories at the beginning of the training phase, they will progressively converge toward being close to each other, known as the consensus problem. Hence, the disagreement strategy seeks to alleviate this issue and boost the performance by keeping two networks diverged within the training epochs or slowing the consensus rate between two networks as the number of epochs increases. \\
\indent Similarly, the lack of distinctiveness is a common problem in self-training SSOD methods employing the EMA strategy, as the weights of the Teacher-Student models become almost identical towards the latter stages of training \cite{DiverseLearner}. To alleviate this problem, we introduce the simple yet effective RD strategy during the SSL stage that encourages the student model to explore further underlying patterns in unlabeled data. It is also motivated by JoCoR \cite{JoCoR} and DML \cite{DeepMutualLearning} methods used for the weakly-supervised learning and knowledge distillation. However, our strategy relies on the representation space to keep the models diverged in contrast with the DML and JoCoR aimed at reducing the diversity between ensemble networks and minimizing the KL divergence between the probabilistic outputs of networks. 
For the proposed RD strategy, we first compute the probability distributions of semantic representations $f_i \in \mathbb{R}^{C \times H \times W}$ from Teacher/Student models using $p_{t}=softmax\left(f_{t}(\bar{x}^{u})\right)$ and $p_{s}=softmax\left(f_{s}(\hat{x}^{u})\right)$, where ${\bar{x}^{u}}$ and $\hat{x}^{u}$ are the weakly-augmented and strongly-augmented samples from unlabeled data, respectively.
Then, the asymmetric KL divergence is computed between the probability distributions based on the supervision of the teacher model as
\vskip -0.85cm
\begin{gather} \label{eq:kl_loss}
\mathcal{L}_{\rm{RD}} = {KL}(p_s(\hat{x}^{u}) || p_t(\bar{x}^{u})). 
\end{gather}
\vskip -0.2cm
\noindent Following that, the learnable parameters for the student model are updated during the SSL stage by
\vskip -0.5cm
\begin{gather} \label{eq:stu_upd_kl}
\theta_{s} \leftarrow \theta_{s}+\xi \frac{\partial(\lambda_u \mathcal{L}_{unsup}-\lambda_d \mathcal{L}_{\rm{RD}})}{\partial \theta_s}          
\end{gather}
\vskip -0.2cm
\noindent where the learning rate is denoted by $\xi$, while $\lambda_u$, and $\lambda_d$ control the contribution of the unsupervised and representation disagreement losses, respectively. Here, $\mathcal{L}_{unsup}$ represents the unsupervised loss of the student model using $\{\hat{x}_{k}^{u}, \tilde{y}_{k}\}_{k=1}^{N_{unsup}}$, in which $\tilde{y}$ denotes pseudo-labels from the teacher model. 
To ensure valid comparisons, we adopted $\mathcal{L}_{\text{unsup}}$ similar to that of our baseline methods \cite{UnbiasedTeacher, UnbiasedTeacher2}, both with and without incorporation of cascade regression.
\section{Empirical Experiments}\label{sec4:exp}
This section includes implementation details of the proposed approach, presents state-of-the-art (SOTA) comparisons using the \texttt{MS-COCO} \cite{COCO} and \texttt{Pascal VOC} \cite{PascalVOC} datasets, and provides systematic ablation analyses. The \texttt{COCO} dataset consists of the \texttt{COCO-standard} (comprising the \texttt{train2017} and \texttt{val2017} splits with $\sim$118k and $\sim$5k labeled images, respectively) and \texttt{COCO-additional} ($\sim$123k unlabeled images) sets. For the \texttt{MS-COCO} dataset, we evaluated our TMR-RD using two experimental settings, including partially-labeled and fully-labeled data. The partially-labeled data setting involves randomly selecting 1\%, 5\%, and 10\% of the \texttt{train2017} split as labeled training data, while the remaining data was treated as unlabeled training data. For the fully-labeled data setting, the entire \texttt{train2017} split and the \texttt{COCO-additional} set are utilized as labeled data and unlabeled data, respectively. The evaluations are performed on the \texttt{val2017} set using the \textit{mean average precision} (mAP) metric. For the \texttt{Pascal VOC} dataset, the training was conducted using \texttt{VOC07-trainval} ($\sim$5k images) and \texttt{VOC12-trainval} ($\sim$11.5k images) as the labeled set and unlabeled set, respectively. The models were then evaluated on the \texttt{VOC07-test} set ($\sim$5k images) using {AP}$_{50}$ and {AP}$_{50:95}$ (denoted as mAP) metrics.
\subsection{Implementation Details} \label{sec:imp_details}
We employ the Faster-RCNN \cite{FasterRCNN} with and without cascade regression implemented in Detectron2 \cite{Detectron2} as our base detection framework. The backbones consist of ResNet-$50$-FPN architecture \cite{ResNet50FPN} initialized with the pre-trained Image-Net \cite{ImageNet} weights. 
The Burn-In and SSL (without classical EMA) stages followed the baselines of Unbiased-Teacher-v1 \cite{UnbiasedTeacher} and Unbiased-Teacher-v2 \cite{UnbiasedTeacher2} with pre-training for $2$k iterations for the \texttt{COCO-standard}, $90$k for the \texttt{COCO-additional}, and $30$k for the \texttt{Pascal VOC}. Then, the trained detectors were duplicated as Teacher-Student models for further training. \\
\indent We experimentally adopted cyclic SSL and TMR stages with $N=4$k iterations dedicated to the SSL and $N'=2$k iterations devoted to the TMR (see Sec.~\ref{sec:ablation}). 
For partially-labeled data setting, the models were trained for $269$k iterations (i.e., $178$k for SSL (as in \cite{UnbiasedTeacher, UnbiasedTeacher2}) \& $89$k for TMR) and $536$k iterations (i.e., $356$k for SSL \& $178$k for TMR) using the base detectors of Faster-RCNN without and with cascade regression, respectively. 
In the fully-labeled data setting, we trained the models with Faster-RCNN without and with cascade regression detectors for $495$k and $900$k iterations, respectively. Specifically, we used $270$k iterations for SSL (as in \cite{UnbiasedTeacher, UnbiasedTeacher2}) and $135$k iterations for TMR using the Faster-RCNN detector, while with cascade regression, we used $540$k iterations for SSL and $270$k iterations for TMR. 
The batch size of $64$ was used in training, where $32$ labeled \& $32$ unlabeled images were randomly selected for each batch.
Following the SSL iterations in baselines \cite{UnbiasedTeacher, UnbiasedTeacher2}, the models were trained on the \texttt{Pascal VOC dataset} for $255$k iterations (i.e., $150$k for SSL \& $75$k for TMR) using the base detector of Faster-RCNN. For Faster-RCNN with cascade regression, the models were trained for $480$k iterations (i.e., $300$k for SSL \& $150$k for TMR) using a batch size of $32$ ($16$ labeled images and $16$ unlabelled images). \\
\indent The implementations were performed on $16$ synchronized Nvidia Tesla V$100$ GPUs with $16$GB RAM. 
The cascade regression utilizes three detection stages with IoU thresholds $\tau=\{0.5, 0.6, 0.7\}$ for generating high-quality pseudo-labels. The loss coefficients were set to $\lambda_t=1$, $\lambda_s=4$, $\lambda_u=4$, while the RD loss coefficient was set to $\lambda_d=0.5$ or $\lambda_d=1$ using the base detector of Faster-RCNN without or with cascade regression, respectively. Moreover, the semantic representations were extracted from the last two levels of the backbones, i.e., $f_i=[\rm{conv4\_x}, \rm{conv5\_x}]$ (typically responsible for capturing complex patterns and representations that are relevant to final predictions). The optimizer, learning rates, data augmentations, and other hyperparameters were applied similarly to those in the baselines \cite{UnbiasedTeacher, UnbiasedTeacher2}.
\begin{table*}[!h]
\captionsetup{font=small}
\caption{
State-of-the-art comparison on the \texttt{MS-COCO} dataset under the partially-labeled data (\texttt{COCO-standard}) and fully-labeled data (\texttt{COCO-additional}) settings. The proposed approach, referred to as TMR-RD-v1 and TMR-RD-v2, is integrated into \cite{UnbiasedTeacher} and \cite{UnbiasedTeacher2}, respectively, with base detectors shown as Faster-RCNN with or without cascade regression.
} 
\vskip -0.3cm
\resizebox{\textwidth}{!}{
\centering 
\begin{tabular}{c | c c c | c} 
\hline \hline
\small {Methods} & {1\% labeled samples} & {5\% labeled samples} & {10\% labeled samples} & {Fully-labeled samples}  \\ \hline

\cellcolor{teal!10}Supervised  & \cellcolor{teal!10}9.05  & \cellcolor{teal!10}18.47 & \cellcolor{teal!10}23.86 & \cellcolor{teal!10}40.2 \\

Supervised (with Cascade regression) & 10.86 & 19.44 & 25.18 & 42.1 \\ \hline

\cellcolor{teal!10}ISMT \cite{ISMT}                              & \cellcolor{teal!10}18.88 & \cellcolor{teal!10}26.37 & \cellcolor{teal!10}30.53 & \cellcolor{teal!10}39.6 \\

Instant-Teaching \cite{InstantTeaching}       & 18.05 & 26.75 & 30.40 & 40.2  \\ 

\cellcolor{teal!10}CAPL \cite{CertaintyPseudo}                   & \cellcolor{teal!10}19.02 & \cellcolor{teal!10}28.40 & \cellcolor{teal!10}32.23 & \cellcolor{teal!10}43.3 \\ 

SED \cite{SED}                                &   -   & 29.01 & 34.02 & 41.5 \\ 

\cellcolor{teal!10}Humble-Teacher \cite{HumbleTeacher}           & \cellcolor{teal!10}16.96 & \cellcolor{teal!10}27.70 & \cellcolor{teal!10}31.61 & \cellcolor{teal!10}42.4  \\ 

Soft-Teacher \cite{SoftTeacher}               & 20.46 & 30.74 & 34.04 & 44.5 \\ 

\cellcolor{teal!10}MA-GCP \cite{MA-GCP}      & \cellcolor{teal!10}21.30 & \cellcolor{teal!10}31.67 & \cellcolor{teal!10}35.02 & \cellcolor{teal!10}45.9 \\ 

Unbiased-Teacher-v1 \cite{UnbiasedTeacher}       & 20.75 & 28.27 & 31.50 & 41.3  \\ 

\cellcolor{teal!10}{Unbiased-Teacher-v2 \cite{UnbiasedTeacher2}} & \cellcolor{teal!10}{25.40} & \cellcolor{teal!10}{31.85} & \cellcolor{teal!10}{35.08} & \cellcolor{teal!10}{44.8}  \\

Active-Teacher \cite{ActiveTeacher}           & 22.20 & 30.07 & 32.58 & - \\ 

\cellcolor{teal!10}Diverse-Learner \cite{DiverseLearner}         & \cellcolor{teal!10}23.72 & \cellcolor{teal!10}31.92 & \cellcolor{teal!10}34.61 & \cellcolor{teal!10}44.8 \\ 

{Pseudo-Polish \cite{Pseudo-Polish}} & {23.55} & {32.10} & {35.30} & {-}  \\ 

\cellcolor{teal!10}{De-biased Teacher \cite{De-biased-Teacher}} & \cellcolor{teal!10}{22.50} & \cellcolor{teal!10}{32.10} & \cellcolor{teal!10}{35.50} & \cellcolor{teal!10}{44.7}  \\ \hline
{TMR-RD-v1 (baseline \cite{UnbiasedTeacher}: Faster-RCNN)} & {22.23} & {30.60} & {33.92} & {43.2} \\

\cellcolor{teal!10}{TMR-RD-v2 (baseline \cite{UnbiasedTeacher2}: Faster-RCNN)} & \cellcolor{teal!10}{26.91} & \cellcolor{teal!10}{34.37} & \cellcolor{teal!10}{37.74} & \cellcolor{teal!10}{46.9} \\ \hdashline

{TMR-RD-v1 (baseline \cite{UnbiasedTeacher} + cascade regression)} & {24.39} & {33.41} & {36.87} & {46.6} \\

\cellcolor{teal!10}{TMR-RD-v2 (baseline \cite{UnbiasedTeacher2} + cascade regression)} & \cellcolor{teal!10}{29.16} & \cellcolor{teal!10}{37.58} & \cellcolor{teal!10}{40.93} & \cellcolor{teal!10}{50.4} \\
\hline \hline
\end{tabular}} \label{table:sota}
\vskip -0.4cm
\end{table*}
\begin{table}[h]
  \vspace{0pt}
  \captionsetup{font=small}
\caption{
State-of-the-art comparison on the \texttt{Pascal VOC}.
} 
\vskip -0.3cm
  \centering
  \resizebox{\linewidth}{!}{%
  \begin{tabular}{c|c|c}
    \hline \hline
    Method & $\rm{AP_{50}}$ & $\rm{AP_{50:95}}$ \\ \hline 
    \cellcolor{teal!10}ISMT \cite{ISMT} & \cellcolor{teal!10}77.23 & \cellcolor{teal!10}46.23 \\ 

    Unbiased-Teacher-v1 \cite{UnbiasedTeacher} & 77.37 & 48.69 \\
    
    \cellcolor{teal!10}Instant-Teaching \cite{InstantTeaching} & \cellcolor{teal!10}78.30 & \cellcolor{teal!10}48.70 \\ 

    Humble-Teacher \cite{HumbleTeacher} & 80.94 & 53.04 \\ 

    \cellcolor{teal!10}Soft-Teacher \cite{SoftTeacher} & \cellcolor{teal!10}80.32 & \cellcolor{teal!10}- \\  

    CAPL \cite{CertaintyPseudo} & 79.0 & 54.60 \\ 

    \cellcolor{teal!10}SED \cite{SED} & \cellcolor{teal!10}80.60 & \cellcolor{teal!10}- \\ 

    MA-GCP \cite{MA-GCP} \cite{MA-GCP} & 81.72 & - \\ 

    \cellcolor{teal!10}Unbiased-Teacher-v2 \cite{UnbiasedTeacher2} & \cellcolor{teal!10}81.29 & \cellcolor{teal!10}56.87 \\ 

    Pseudo-Polish \cite{Pseudo-Polish} & 82.50 & 52.40 \\ 

    \cellcolor{teal!10}De-biased Teacher \cite{De-biased-Teacher} & \cellcolor{teal!10}81.50 & \cellcolor{teal!10}- \\ \hline
    
    TMR-RD-v1 (baseline \cite{UnbiasedTeacher}: Faster-RCNN) & {79.83} & {51.96} \\ 
    
    \cellcolor{teal!10}TMR-RD-v2 (baseline \cite{UnbiasedTeacher2}: Faster-RCNN) & \cellcolor{teal!10}{83.66} & \cellcolor{teal!10}{60.23} \\ \hdashline
    
    TMR-RD-v1 (baseline \cite{UnbiasedTeacher} + cascade regression) & {82.68} & {55.08} \\ 
    
    \cellcolor{teal!10}TMR-RD-v2 (baseline \cite{UnbiasedTeacher2} + cascade regression) & \cellcolor{teal!10}{85.92} & \cellcolor{teal!10}{63.48} \\
    \hline \hline
  \end{tabular}}
  \label{tab:sot_voc}
  \vskip -0.6cm
\end{table}
\subsection{State-of-the-art Comparison}
In this section, we compare the proposed TMR-RD approach integrated into Unbiased-Teacher-v1 \cite{UnbiasedTeacher} (referred to as TMR-RD-v1) and Unbiased-Teacher-v2 \cite{UnbiasedTeacher2} (referred to as TMR-RD-v2) with 13 recent SSOD methods, namely 
ISMT \cite{ISMT}, 
Instant-Teaching \cite{InstantTeaching}, 
CAPL \cite{CertaintyPseudo}, 
SED \cite{SED}, 
Humble-Teacher \cite{HumbleTeacher}, 
Soft-Teacher \cite{SoftTeacher}, 
MA-GCP \cite{MA-GCP}, 
Unbiased-Teacher-v1 \cite{UnbiasedTeacher} (baseline 1), 
Unbiased-Teacher-v2 \cite{UnbiasedTeacher2} (baseline 2), 
Active-Teacher \cite{ActiveTeacher}, 
Diverse-Learner \cite{DiverseLearner}, 
Pseudo-Polish \cite{Pseudo-Polish}, and
De-biased Teacher \cite{De-biased-Teacher}. 
Table~\ref{table:sota} and Table~\ref{tab:sot_voc} provide the results on the \texttt{MS-COCO} and \texttt{Pascal VOC} datasets, respectively.
To ensure fair comparisons, we compare the two versions of the proposed TMR-RD using the Faster-RCNN against SOTA methods. Additionally, we examine improvements resulting from cascade regression compared to their baselines, aimed at generating more reliable pseudo-labels. We highlight the generalization capability of our approach through extensive empirical experiments, with and without cascade regression.

\textbf{\texttt{COCO-standard}}: 
First, we compare the proposed approach with existing SOTA methods partially trained on the \texttt{COCO-standard} set. As shown in Table~\ref{table:sota}, our TMR-RD-v2 considerably outperforms the Diverse-Learner \cite{DiverseLearner}, Pseudo-Polish \cite{Pseudo-Polish}, and De-biased Teacher \cite{De-biased-Teacher} by average margins of 2.92, 2.69, and 2.97 mAP across all labeling ratios, respectively. In particular, our TMR-RD demonstrates superior performance over the Unbiased-Teacher-v1 \cite{UnbiasedTeacher} and Unbiased-Teacher-v2 \cite{UnbiasedTeacher2} baselines by an average of 2.07 and 2.23 mAP, respectively. Furthermore, integrating the proposed approach with cascade regression surpasses the performance of both TMR-RD-v1 and TMR-RD-v2 by average margins of 2.64 and 2.88 mAP, respectively. 

\textbf{\texttt{COCO-additional}}:
We then evaluate the performance of the proposed approach fully trained on supervised data to determine the extent of further improvement by incorporating additional unlabeled data. According to Table~\ref{table:sota}, our TMR-RD-v2 provides an mAP improvement of 2.1, 2.1, and 2.2 compared to the Unbiased-Teacher-v2 \cite{UnbiasedTeacher2}, Diverse-Learner \cite{DiverseLearner}, and De-biased Teacher \cite{De-biased-Teacher}, respectively. Moreover, the proposed TMR-RD-v1 and TMR-RD-v2 outperform their corresponding supervised models by 3 and 6.7 mAP, respectively. Further, the utilization of cascade regression demonstrates superior performance, surpassing the mAP of TMR-RD-v1 and TMR-RD-v2 by 3.4 and 3.5, respectively.
These comparisons highlight the effectiveness of the proposed approach with accessible large amounts of labeled and unlabeled data, in addition to partially labeled datasets where labeled data is limited. 

\textbf{\texttt{Pascal VOC}}: 
At last, the proposed TMR-RD demonstrates its superiority over existing SSOD methods, as shown in Table~\ref{tab:sot_voc}. For example, TMR-RD-v2 outperforms De-biased Teacher \cite{De-biased-Teacher}, MA-GCP \cite{MA-GCP}, and Pseudo-Polish \cite{Pseudo-Polish} by margins of up to 2.16, 1.94, and 1.16 in the AP$_{50}$ metric, respectively. Also, it surpasses Unbiased-Teacher-v1 \cite{UnbiasedTeacher} and Unbiased-Teacher-v2 \cite{UnbiasedTeacher2} by 3.36 and 3.27 mAP, respectively. Additionally, the integration of cascade regression yields superior results, surpassing TMR-RD-v2 and TMR-RD-v1, leading to improvements of 3.25 and 3.12 mAP, respectively.
\subsection{Ablation Analysis} \label{sec:ablation}
A systematic ablation analysis of the proposed approach, integrated into the Unbiased-Teacher-v1 \cite{UnbiasedTeacher} and Unbiased-Teacher-v2 \cite{UnbiasedTeacher2}, has been conducted using a 5\% labeled \texttt{COCO-standard} and \texttt{Pascal VOC} datasets. The analysis includes $12$ configurations (see Table~\ref{tab:ab_coco} \& Table~\ref{tab:ab_voc}), including: 
i) baseline \cite{UnbiasedTeacher} with the Faster-RCNN and classical EMA (A1), 
ii) A1 equipped with TMR stage (A11), 
iii) A11 equipped with RD strategy (A12), 
iv) baseline \cite{UnbiasedTeacher2} with the Faster-RCNN and classical EMA (A2), 
v) A2 equipped with TMR stage (A21), 
vi) A21 equipped with RD strategy (A22),
vii) baseline \cite{UnbiasedTeacher} equipped with cascade regression and classical EMA (A3), 
viii) A3 equipped with TMR stage (A31), 
ix) A31 equipped with RD strategy (A32), 
x) baseline \cite{UnbiasedTeacher2} equipped with cascade regression and classical EMA (A4), 
xi) A4 equipped with TMR stage (A41), and
xii) A41 equipped with RD strategy (A42). 
These analyses validate the effectiveness of our TMR stage and RD strategy.

\begin{figure}[t]
  \vspace{0pt}
\noindent\begin{minipage}[b]{0.49\linewidth}
    \centering
    \includegraphics[width=\linewidth]{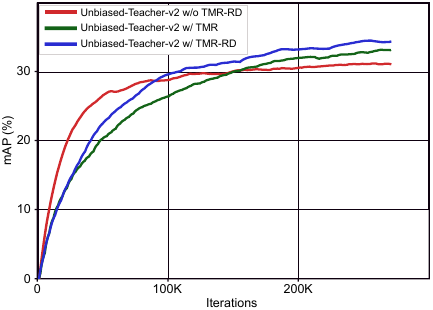}
    \text{(a)}
  \end{minipage}
  \hfill
  \begin{minipage}[b]{0.49\linewidth}
    \centering
    \includegraphics[width=\linewidth]{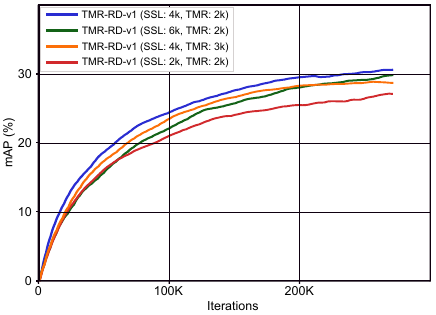}
    \text{(b)}
  \end{minipage}
  \vskip -0.3cm
  \caption{Ablation study of mAP improvement in the case of \texttt{COCO-standard} 5\% labeled data. (a): Integration of the TMR stage and RD strategy into Unbiased-Teacher-v2 \cite{UnbiasedTeacher2} for effectiveness evaluation. (b): Exploration of the impact of varying SSL and TMR stages using Unbiased-Teacher-v1 \cite{UnbiasedTeacher}. Best viewed in color and zoom-in.
  }
  \label{fig:curve}
  \vskip -0.4cm
\end{figure}

\textbf{Effect of TMR stage}: 
We first ablate the impact of the proposed TMR stage integrated into both baselines with or without cascade regression. According to the results of A11, A21, A31, and A41 in Table~\ref{tab:ab_coco} and Table~\ref{tab:ab_voc}, the baselines of \cite{UnbiasedTeacher2} and \cite{UnbiasedTeacher} with the Faster-RCNN are notably outperformed by the proposed TMR stage, with average mAP improvements of 2.12 and 2.03, respectively. Moreover, incorporating our TMR stage considerably enhances the performance of these baselines by 2.69 and 2.47 mAP, respectively, when employing the base detector with cascade regression. 
These results further demonstrate the versatility of our method and its ability to improve overall performance and generalization of existing methods.
We also validated the efficacy of the TMR stage by visually analyzing the mAP improvement of baseline \cite{UnbiasedTeacher2} in Fig.~\ref{fig:curve}(a). 
While our model (shown in green) was trained with fewer SSL steps (For comparability purposes considering our cyclic SSL and TMR stages), our method effectively refines the models to transfer reliable knowledge while preventing them from misleading information and converging too early (unlike the baseline \cite{UnbiasedTeacher2} shown in red). 

\begin{table}[t]
  \vspace{0pt}
    \caption{
Ablation study of the proposed TMR-RD approach on the \texttt{COCO-standard} set. Classical EMA and cascade regression are denoted as cEMA and c-regress, respectively.
} 
\vskip -0.3cm
  \centering
  \resizebox{1\linewidth}{!}{
  \begin{tabular}{c | c c c c c c | c}
    \hline \hline
     Abl. & A1 & A2 & cEMA & TMR & RD & c-regress. & $\rm{mAP}$ \\ \hline 

     \cellcolor{teal!10}A1 \cite{UnbiasedTeacher} & \cellcolor{teal!10}\checkmark & \cellcolor{teal!10}$\times$ & \cellcolor{teal!10}\checkmark & \cellcolor{teal!10}$\times$ & \cellcolor{teal!10}$\times$ & \cellcolor{teal!10}$\times$ & \cellcolor{teal!10}{28.27} \\

     A11 & \checkmark & $\times$ & $\times$ & \checkmark & $\times$ & $\times$ & 29.93 \\

     \cellcolor{teal!10}A12 & \cellcolor{teal!10}\checkmark & \cellcolor{teal!10}$\times$ & \cellcolor{teal!10}$\times$ & \cellcolor{teal!10}\checkmark & \cellcolor{teal!10}\checkmark & \cellcolor{teal!10}$\times$ & \cellcolor{teal!10}30.60 \\ \hline

     A2 \cite{UnbiasedTeacher2} & $\times$ & \checkmark & \checkmark & $\times$ & $\times$ & $\times$ & 31.85 \\

     \cellcolor{teal!10}A21 & \cellcolor{teal!10}$\times$ & \cellcolor{teal!10}\checkmark & \cellcolor{teal!10}$\times$ & \cellcolor{teal!10}\checkmark & \cellcolor{teal!10}$\times$ & \cellcolor{teal!10}$\times$ & \cellcolor{teal!10}33.63 \\

     A22 & $\times$ & \checkmark & $\times$ & \checkmark & \checkmark & $\times$ & 34.37 \\ \hline

     \cellcolor{teal!10}A3 & \cellcolor{teal!10}\checkmark & \cellcolor{teal!10}$\times$ & \cellcolor{teal!10}\checkmark & \cellcolor{teal!10}$\times$ & \cellcolor{teal!10}$\times$ & \cellcolor{teal!10}\checkmark & \cellcolor{teal!10}\cellcolor{teal!10}{30.19} \\

     A31 & \checkmark & $\times$ & $\times$ & \checkmark & $\times$ & \checkmark & {32.66} \\

     \cellcolor{teal!10}A32 & \cellcolor{teal!10}\checkmark & \cellcolor{teal!10}$\times$ & \cellcolor{teal!10}$\times$ & \cellcolor{teal!10}\checkmark & \cellcolor{teal!10}\checkmark & \cellcolor{teal!10}\checkmark & \cellcolor{teal!10}\cellcolor{teal!10}{33.41} \\ \hline

     A4 & $\times$ & \checkmark & \checkmark & $\times$ & $\times$ & \checkmark & {34.05} \\

     \cellcolor{teal!10}A41 & \cellcolor{teal!10}$\times$ & \cellcolor{teal!10}\checkmark & \cellcolor{teal!10}$\times$ & \cellcolor{teal!10}\checkmark & \cellcolor{teal!10}$\times$ & \cellcolor{teal!10}\checkmark & \cellcolor{teal!10}{36.74} \\

     A42 & $\times$ & \checkmark & $\times$ & \checkmark & \checkmark & \checkmark & {37.58} \\  \hline \hline
  \end{tabular}}
  \label{tab:ab_coco}
\vskip -0.3cm
\end{table}
\begin{table}[t]
  \caption{
Ablation study of the proposed TMR-RD approach on the \texttt{Pascal VOC} dataset.
} 
\vskip -0.3cm
  \centering
  \resizebox{1\linewidth}{!}{
  \begin{tabular}{c | c c c c c | c | c}
    \hline \hline
    Abl. & A1 & A2 & cEMA & TMR & RD & $\rm{AP_{50}}$ & $\rm{AP_{50:95}}$ \\ \hline 

     \cellcolor{teal!10}A1 \cite{UnbiasedTeacher} & \cellcolor{teal!10}\checkmark & \cellcolor{teal!10}$\times$ & \cellcolor{teal!10}\checkmark & \cellcolor{teal!10}$\times$ & \cellcolor{teal!10}$\times$ & \cellcolor{teal!10}77.37 & \cellcolor{teal!10}48.69  \\

     A11 & \checkmark & $\times$ & $\times$ & \checkmark & $\times$ & 79.21 & 51.09  \\

     \cellcolor{teal!10}A12 & \cellcolor{teal!10}\checkmark & \cellcolor{teal!10}$\times$ & \cellcolor{teal!10}$\times$ & \cellcolor{teal!10}\checkmark & \cellcolor{teal!10}\checkmark & \cellcolor{teal!10}79.83 & \cellcolor{teal!10}51.96  \\ \hline

     A2 \cite{UnbiasedTeacher2} & $\times$  & \checkmark & \checkmark & $\times$ & $\times$ & 81.29 & 56.87  \\

     \cellcolor{teal!10}A21 & \cellcolor{teal!10}$\times$ & \cellcolor{teal!10}\checkmark & \cellcolor{teal!10}$\times$ & \cellcolor{teal!10}\checkmark & \cellcolor{teal!10}$\times$ & \cellcolor{teal!10}82.94 & \cellcolor{teal!10}59.32  \\

     A22 & $\times$ & \checkmark & $\times$ & \checkmark & \checkmark & 83.66 & 60.23  \\ \hline \hline
  \end{tabular}}
  \label{tab:ab_voc}
\vskip -0.5cm
\end{table}
\textbf{Effect of RD Strategy}: 
The results of experiments A12, A22, A32, and A42 provide compelling support for the usefulness of the proposed RD strategy. These results demonstrate that the performance of \cite{UnbiasedTeacher2} and \cite{UnbiasedTeacher} baselines utilizing the Faster-RCNN detector are surpassed by employing the proposed RD strategy with average mAP improvements of 0.82 and 0.77, respectively. Notably, the integration of this strategy led to satisfactory effects in the performance of the baseline \cite{UnbiasedTeacher2} as shown in Fig.~\ref{fig:curve}(a). It validates the complementary effect of this strategy and demonstrates how it promotes the mutual learning of Teacher-Student models by reinforcing model distinctiveness.

\textbf{Effect of SSL and TMR iterations}: We conducted additional ablations, as illustrated in Fig.~\ref{fig:curve}(b), to analyze the impacts of various SSL and TMR iterations on model performance using baseline \cite{UnbiasedTeacher}. 
The best result (blue curve) was achieved when employing cyclic SSL and TMR stages with $N=4$k and $N'=2$k iterations.
The red curve highlights the necessity for models to undergo sufficient evolution during the SSL stage before each subsequent TMR refinement while increasing the SSL stage (green curve) can result in a delay in model refinement and slight performance degradation, as the teacher model is expected to generate more accurate predictions than the student model \cite{HumbleTeacher}. Also, increasing the number of TMR iterations with partially labeled data can impede the subsequent joint evolution of models using unlabeled data (orange curve). 
\noindent
\section{Conclusion} \label{sec5:conclusion}
In this paper, we presented a novel training-based model refinement stage and a representation disagreement strategy for existing SSOD frameworks. To address the limitations of the classical EMA strategy, our proposed model refinement stage learns lightweight scaling operation parameters to dynamically refine the weights of Teacher-Student models, ensuring that overfitting and forgetting learned patterns from unlabeled data are avoided. Moreover, we introduce a simple yet effective representation disagreement strategy to alleviate the consensus of Teacher-Student models that arise with the progression of model training. This strategy promotes model distinctiveness and prevents the models from converging too early, encouraging the student model to explore more underlying patterns in unlabeled data. Our approach can be integrated into existing SSOD methods to improve performance and generalization, transfer reliable knowledge from the student to the teacher, and prune noisy student weights.
Extensive experiments demonstrate notable performance improvements with different baseline detectors, as well as the potential to integrate the proposed approach into existing SSOD methods. \\
\vskip -0.15cm
\noindent\textbf{Acknowledgments.} 
We thank the fRI Research - MPB Ecology Program for funding and the Digital Research Alliance of Canada for providing the computational resources for our experiments.
\vskip -0.1cm
{\small
\bibliographystyle{ieee_fullname}
\bibliography{egbib}
}

\end{document}